\documentclass[conference]{IEEEtran}
\IEEEoverridecommandlockouts
\usepackage{cite}
\usepackage{amsmath,amssymb,amsfonts}
\usepackage{algorithmic}
\usepackage{graphicx}
\usepackage{xspace}
\usepackage{textcomp}
\usepackage{xcolor}
\usepackage{placeins}
\usepackage{capt-of}
\usepackage{booktabs}
\usepackage{multirow}
\usepackage{dblfloatfix}
\usepackage{afterpage}
\usepackage{etoolbox}
\usepackage{needspace}
\usepackage{placeins}
\usepackage{cuted}
\makeatletter
\patchcmd{\@makecaption}
  {\normalfont\footnotesize\scshape}
  {\normalfont\footnotesize}
  {}{}
\makeatother

\newcommand{\sys}{LiteNWM\xspace}

\def\BibTeX{{\rm B\kern-.05em{\sc i\kern-.025em b}\kern-.08em
T\kern-.1667em\lower.7ex\hbox{E}\kern-.125emX}}
\makeatletter
\newsavebox{\DWMTableNatural}
\newsavebox{\DWMTableOriginal}
\newsavebox{\DWMTableNew}
\newlength{\DWMTableWidth}
\newlength{\DWMTableBaseline}
\newcommand{\DWMUniformTable}[6]{%
  \begingroup
  #2%
  \setlength{\tabcolsep}{#3}%
  \sbox{\DWMTableNatural}{\begin{tabular}{#5}#6\end{tabular}}%
  \ifdim#1>0pt
    \sbox{\DWMTableOriginal}{%
      \resizebox{#1}{!}{\usebox{\DWMTableNatural}}}%
  \else
    \sbox{\DWMTableOriginal}{\usebox{\DWMTableNatural}}%
  \fi
  \setlength{\DWMTableWidth}{\wd\DWMTableOriginal}%
  \Gscale@div\DWMTableScale{\DWMTableWidth}{\wd\DWMTableNatural}%
  \setlength{\DWMTableBaseline}{\DWMTableScale\baselineskip}%
  \setlength{\arrayrulewidth}{\DWMTableScale\arrayrulewidth}%
  \fontsize{8}{\DWMTableBaseline}\selectfont
  \setlength{\tabcolsep}{0pt}%
  \sbox{\DWMTableNew}{\begin{tabular}{#5}#6\end{tabular}}%
  \setlength{\tabcolsep}{\dimexpr\DWMTableWidth-\wd\DWMTableNew\relax}%
  \divide\tabcolsep by #4\relax
  \divide\tabcolsep by 2\relax
  \ifdim\tabcolsep<0pt
    \PackageError{DWMTable}{Table text exceeds original width at 8 pt}%
      {The column content cannot fit without changing the table width.}%
  \fi
  \sbox{\DWMTableNew}{\begin{tabular}{#5}#6\end{tabular}}%
  \typeout{DWM TABLE: width=\the\DWMTableWidth, sep=\the\tabcolsep,
    old-height=\the\ht\DWMTableOriginal+\the\dp\DWMTableOriginal,
    new-height=\the\ht\DWMTableNew+\the\dp\DWMTableNew}%
  \raisebox{0pt}[\ht\DWMTableOriginal][\dp\DWMTableOriginal]{%
    \usebox{\DWMTableNew}}%
  \endgroup
}
\makeatother

\newcommand{\MainResultsTable}{%
    \centering
    \refstepcounter{table}
    \label{tab:main_results}

    {\footnotesize
    \noindent
    \parbox{\linewidth}{%
        \raggedright
        \textbf{TABLE~\thetable}\enspace
        Cross-domain offline trajectory evaluation at $H=4$.
        For $\dagger$ baselines, $\pm$ denotes the 95\% 
        trajectory-cluster-robust confidence-interval half-width;
        otherwise, it denotes the standard deviation across seeds.%
        \par
    }}%
    \vspace{4pt}%
  
  \setlength{\tabcolsep}{3.2pt}
  \scriptsize
  \begingroup
\fontsize{7.2}{8.6}\selectfont
\renewcommand{\arraystretch}{1.05}
\setlength{\tabcolsep}{7pt}
\setlength{\heavyrulewidth}{0.5pt}
\setlength{\lightrulewidth}{0.4pt}
\setlength{\cmidrulewidth}{0.3pt}
\begingroup
\setlength{\tabcolsep}{2.6pt}
\renewcommand{\arraystretch}{1.04}
\resizebox{\linewidth}{!}{%
\begin{tabular}{@{}lccccccccc@{}}
  \toprule
  \multirow{2}{*}{Method}
  & \multicolumn{3}{c}{RECON}
  & \multicolumn{3}{c}{SCAND}
  & \multicolumn{3}{c}{SACSoN} \\
  \cmidrule(lr){2-4}
  \cmidrule(lr){5-7}
  \cmidrule(lr){8-10}
  & ATE & RPE & SR@1$\uparrow$
  & ATE & RPE & SR@1$\uparrow$
  & ATE & RPE & SR@1$\uparrow$ \\
  \midrule

  GNM$^{\dagger}$
  & $0.42\!\pm\!0.05$ & $0.20\!\pm\!0.03$ & 86.0
  & $0.44\!\pm\!0.09$ & $0.21\!\pm\!0.03$ & 89.0
  & $0.91\!\pm\!0.12$ & $0.45\!\pm\!0.06$ & 45.0 \\

  ViNT$^{\dagger}$
  & $0.40\!\pm\!0.05$ & $0.20\!\pm\!0.03$ & 93.0
  & $0.47\!\pm\!0.11$ & $0.22\!\pm\!0.04$ & 81.0
  & $0.77\!\pm\!0.11$ & $0.41\!\pm\!0.06$ & 53.0 \\

  NoMaD
  & $0.47\!\pm\!0.01$ & $0.23\!\pm\!0.00$ & 82.3
  & $0.44\!\pm\!0.02$ & $0.22\!\pm\!0.01$ & 86.7
  & $0.78\!\pm\!0.01$ & $0.43\!\pm\!0.00$ & 60.7 \\

  \midrule
  NWM-S$^{\ddagger}$
  & $0.52\!\pm\!0.01$ & $0.22\!\pm\!0.00$ & 81.5
  & $0.58\!\pm\!0.01$ & $0.24\!\pm\!0.01$ & 77.3
  & $1.20\!\pm\!0.03$ & $0.57\!\pm\!0.01$ & 27.0 \\

  NWM-XL$^{\ddagger}$
  & $0.53\!\pm\!0.01$ & $0.22\!\pm\!0.00$ & 81.0
  & $0.66\!\pm\!0.00$ & $0.26\!\pm\!0.00$ & 71.3
  & $1.07\!\pm\!0.02$ & $0.54\!\pm\!0.01$ & 39.0 \\

  NoMaD+NWM-S
  & $0.44\!\pm\!0.01$ & $0.21\!\pm\!0.00$ & 85.8
  & $0.44\!\pm\!0.00$ & $0.22\!\pm\!0.00$ & 86.5
  & $0.77\!\pm\!0.02$ & $0.42\!\pm\!0.01$ & 57.5 \\

  NoMaD+NWM-XL
  & $0.44\!\pm\!0.01$ & $0.21\!\pm\!0.01$ & 87.3
  & $0.45\!\pm\!0.02$ & $0.22\!\pm\!0.00$ & 86.0
  & $0.77\!\pm\!0.03$ & $0.42\!\pm\!0.01$ & 60.3 \\

  DINO-WM
  & $0.42\!\pm\!0.01$ & $0.20\!\pm\!0.00$ & 89.3
  & $0.58\!\pm\!0.02$ & $0.26\!\pm\!0.01$ & 73.8
  & $0.74\!\pm\!0.02$ & $0.40\!\pm\!0.01$ & 60.8 \\

  \midrule

  \textbf{\sys (Ours)}
  & $\mathbf{0.33\!\pm\!0.01}$ & $\mathbf{0.17\!\pm\!0.00}$ & \textbf{93.8}
  & $\mathbf{0.38\!\pm\!0.01}$ & $\mathbf{0.20\!\pm\!0.00}$ & \textbf{89.5}
  & $\mathbf{0.66\!\pm\!0.00}$ & $\mathbf{0.37\!\pm\!0.00}$ & \textbf{67.3} \\

  \bottomrule
\end{tabular}%
}
\endgroup
\endgroup}
\newcommand{\AuxResultsTables}{%
  \centering  
\begin{minipage}[t]{0.48\textwidth}

\refstepcounter{table}
\label{tab:ablation}

{\footnotesize
\noindent
\parbox{\linewidth}{%
\raggedright
\textbf{TABLE~\thetable}\enspace
Predictive-feature ablation and correspondence diagnostic.
Three-domain development macro with fixed $K{=}16$, lower is better.
The first four rows use separately trained matched scorers;
the last averages two prespecified candidate--future mismatches
applied to the frozen scorer.
\par
}}

\vspace{0.45em}

\centering
\setlength{\tabcolsep}{5pt}
\small
  \DWMUniformTable{\linewidth}{\small}{5pt}{4}{l|ccc}{
      \hline
      Variant & ATE & RPE & FDE \\
      \hline
      No predicted future & 0.4790 & 0.2596 & 0.6743 \\
      Static DINO repeat  & 0.4803 & 0.2610 & 0.6763 \\
      Predicted H4 only   & 0.4662 & 0.2529 & 0.6500 \\
      \textbf{Predicted H1--H4}
        & \textbf{0.4431} & \textbf{0.2316} & \textbf{0.6440} \\
      \hline
      H1--H4, mismatched  & 0.4925 & 0.2672 & 0.6996 \\
      \hline
  }
\end{minipage}
  \hfill
\begin{minipage}[t]{0.48\textwidth}

\refstepcounter{table}
\label{tab:efficiency}

{\footnotesize
\noindent
\parbox{\linewidth}{%
\raggedright
\textbf{TABLE~\thetable}\enspace
System performance on RTX 5090. 
Per-state wall time, GPU energy,
and peak VRAM are measured throughout single action; active parameters are
the sum of whole-workflow modules.
\par
}}

\vspace{0.45em}

\centering
\setlength{\tabcolsep}{3.5pt}
\scriptsize
\DWMUniformTable{\linewidth}{\scriptsize}{3.5pt}{5}{l|rrrr}{
    \hline
    Method & Wall (s) & Energy (Wh) & VRAM (MiB) & Active (M) \\
    \hline
    GNM          & 0.076  & 0.0011  & 1,252  & 8.66 \\
    ViNT         & 0.088  & 0.0014  & 1,658  & 29.66 \\
    NoMaD        & 0.121  & 0.0021  & 791    & 19.05 \\
    DINO-WM      & 2.811  & 0.3560  & 3,520  & 41.30 \\
    NWM-S        & 11.635 & 1.5929 & 8,344  & 49.23 \\
    NWM-XL       & 88.449 & 14.070 & 13,776 & 1012 \\
    NoMaD+NWM-S  & 11.653 & 1.5929 & 8,344 & 154.40 \\
    NoMaD+NWM-XL & 88.449 & 14.0695 & 13,776 & 1,117.13 \\
    \textbf{\sys} & 0.691 & 0.0277 & 2,459 & 349.04 \\
    \hline
}
\end{minipage}
}
  
\begin{document}

\title{
\sys: Efficient Latent World Models for Onboard Visual Navigation in the Wild
}
\author{
\IEEEauthorblockN{
Linkai Liu$^{1,2,\dagger}$,
Yuntian Zhang$^{1,\dagger}$,
Zhenshan Bing$^{1}$,
Chen Chen$^{3}$,
Lingjuan Lyu$^{3}$
}
\IEEEauthorblockN{
Shangguang Wang$^{4}$,
Mengwei Xu$^{4}$,
and Dongqi Cai$^{1,*}$
}

\vspace{2pt}

\IEEEauthorblockA{
$^{1}$Nanjing University
\qquad
$^{2}$Imperial College London
\qquad
$^{3}$Sony AI
\\
$^{4}$Beijing University of Posts and Telecommunications
}
}
\IEEEaftertitletext{%
    \vspace{-0.5\baselineskip}
    \begin{center}
        \makebox[\textwidth][c]{%
            \includegraphics[width=1.06\textwidth]{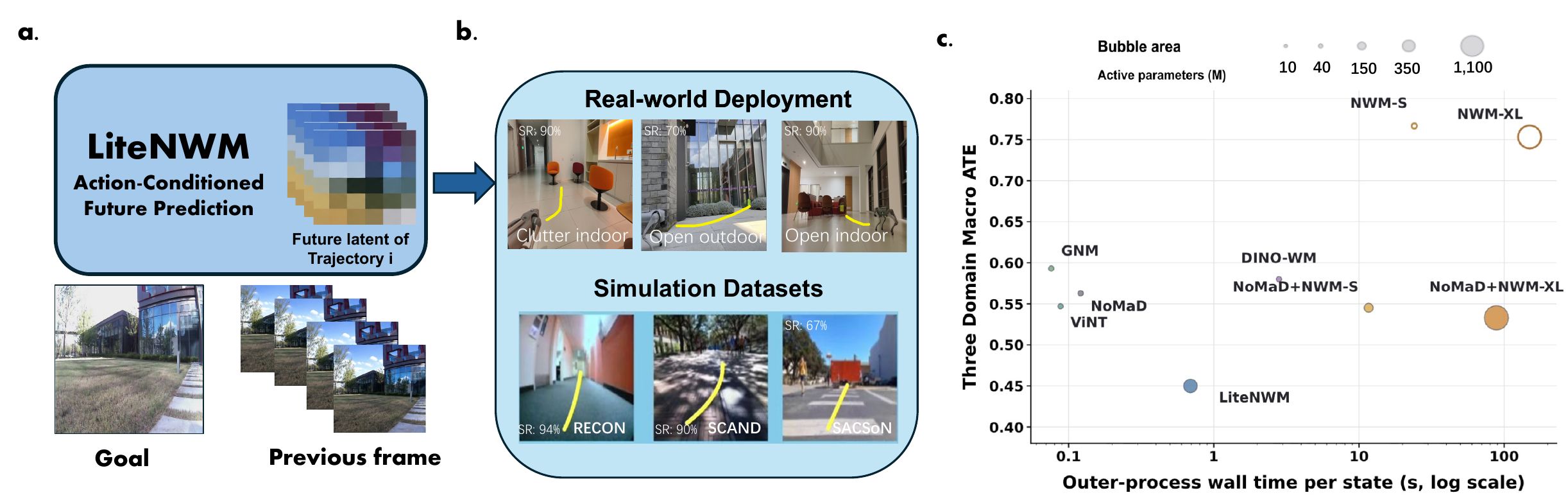}%
        }
        \captionof{figure}{%
            Overview of \sys and main results.
            (a) \sys predicts and scores action-conditioned future trajectories
            from the visual history and goal;
            (b) Closed-loop real-world/simulation execution of the trajectory selected. SR: success rate.;
            (c) Accuracy--efficiency comparison.
        }
        \label{fig:motivation}
    \end{center}
    \vspace{-0.5\baselineskip}%
}

\maketitle
\begin{abstract}
Direct visual navigation policies generate trajectories efficiently but do not explicitly evaluate their future consequences.
Generative navigation world models provide this foresight through visual rollouts, which are costly when evaluating multiple candidates.
We present \sys, a latent navigation world model that shares visual encoding across candidates and jointly predicts their action-conditioned future representations at multiple horizons and a learned scorer uses these predictions to select trajectories.
In offline evaluations on RECON, SCAND, and SACSoN, \sys reduces macro-averaged trajectory error by 17.56\% relative to NoMaD+NWM-XL and achieves a 128.00$\times$ end-to-end speedup on an RTX~5090.
The same evaluator transfers from NoMaD to MBRA without proposer-specific retraining, reducing MBRA's macro-averaged trajectory error by 16.2\%.
In real-robot experiments in unseen indoor and outdoor environments, \sys improves navigation success from 43.3\% to 83.3\% relative to NoMaD.
These results demonstrate that \sys can be deployed for future-aware planning and closed-loop navigation on a physical robot.
\end{abstract}


\section{Introduction}

Goal-conditioned visual navigation enables a robot to reach a visually specified target using egocentric observations~\cite{targetnav}.
Recent generalist navigation policies, such as GNM, ViNT, and NoMaD, learn navigation from large and diverse robot datasets~\cite{gnm,vint,nomad}.
Diffusion-based policies can further generate multiple plausible trajectories for each observation.
However, these policies select trajectories without explicitly evaluating their future consequences, which can lead to suboptimal decisions in challenging environments.

As illustrated in Fig.~\ref{fig:motivation}, Navigation World Models (NWM) address this limitation through future-aware trajectory selection~\cite{nwm}.
Given multiple trajectories proposed by a navigation policy, NWM predicts future visual observations conditioned on each candidate and selects the trajectory based on its predicted outcome.

In our component profile with prediction horizon $H=4$ and $K=16$ candidate trajectories, candidate-conditioned RGB rollout accounts for 97.6\% of the captured NoMaD+NWM-XL computation (Fig.~\ref{fig:runtime_breakdown}). Its end-to-end wall time is 221.97~s per decision state (Sec.~\ref{subsec:efficiency}), motivating a less expensive evaluation path.

We present \sys, a navigation trajectory evaluator designed around the rollout bottleneck identified above. 
Action-conditioned latent prediction already supports navigation without RGB reconstruction.
\sys targets candidate-wise rollout cost through shared visual encoding and joint H1--H4 prediction across candidates. A learned incumbent-relative scorer uses the predicted consequences and proposal-set context to guide trajectory selection. This design combines an efficient prediction path with relative outcome evaluation while keeping the navigation policies frozen.

We evaluate \sys on development cohorts from three navigation datasets and examine its deployment on a physical robot.
\sys improves trajectory accuracy while achieving 128.00$\times$ and 16.86$\times$ end-to-end speedups over NoMaD+NWM-XL and NoMaD+NWM-S, respectively, on an RTX~5090 (Fig.~\ref{fig:motivation}(d)).
The frozen evaluator also transfers to MBRA without proposer-specific retraining.
Ablations attribute the accuracy gain to dynamic multi-horizon latent prediction.
\sys supports closed-loop navigation on a Unitree Go2 EDU in unseen environments.
Our main contributions are:
\begin{itemize}
   \item We identify candidate-wise RGB rollout as the dominant evaluation cost and restructure the evaluation path around shared encoding and joint candidate-future prediction.

  \item We combine pose-conditioned prediction with incumbent-relative scoring to compare candidate consequences using proposal-set context.

  \item We validate accuracy and end-to-end efficiency on matched initial proposal pools, achieving a 128.00$\times$ speedup over NoMaD+NWM-XL on an RTX~5090, and demonstrate cross-proposer transfer and real-robot deployment.
\end{itemize}

\section{Related Work}

\subsection{Generalist Navigation Policies and Trajectory Selection}
Learning-based visual navigation has increasingly shifted toward policies trained on heterogeneous robot experience.
GNM introduced a shared goal-conditioned formulation across robot
platforms \cite{gnm}, while ViNT scaled this direction with a Transformer trained on diverse navigation data \cite{vint}.
NoMaD further combines goal masking with a diffusion decoder to represent multimodal navigation behaviors \cite{nomad,diffusionpolicy}.
Proposal generation and trajectory selection become distinct operations when multiple stochastic trajectories are exposed at inference.
NavDP couples diffusion-based trajectory generation with a learned critic for selection \cite{navdp}, while ViLiNT ranks generated trajectories using an embodiment-conditioned path-clearance predictor \cite{vilint}.
\sys keeps trajectory generation fixed and focuses on evaluating the resulting proposals using predicted future evidence.

\subsection{Generative World Models for Navigation}
Visual world models provide such evidence by predicting observations along candidate action sequences.
Pathdreamer generates unobserved panoramic views for look-ahead planning in indoor navigation \cite{pathdreamer}.
Moreover, NWM predicts future RGB observations conditioned on candidate actions and uses the synthesized futures both for standalone planning and for ranking trajectories sampled from an external policy \cite{nwm}.
NWM therefore establishes that future-conditioned trajectory reranking and a modular proposer--evaluator interface are prior art.
The question narrows to whether short-horizon trajectory candidates can instead be compared directly from action-conditioned future latent features rather than reconstructing an RGB future for every candidate.

\begin{figure*}[!t]
    \centering
    \includegraphics[width=0.94\textwidth]{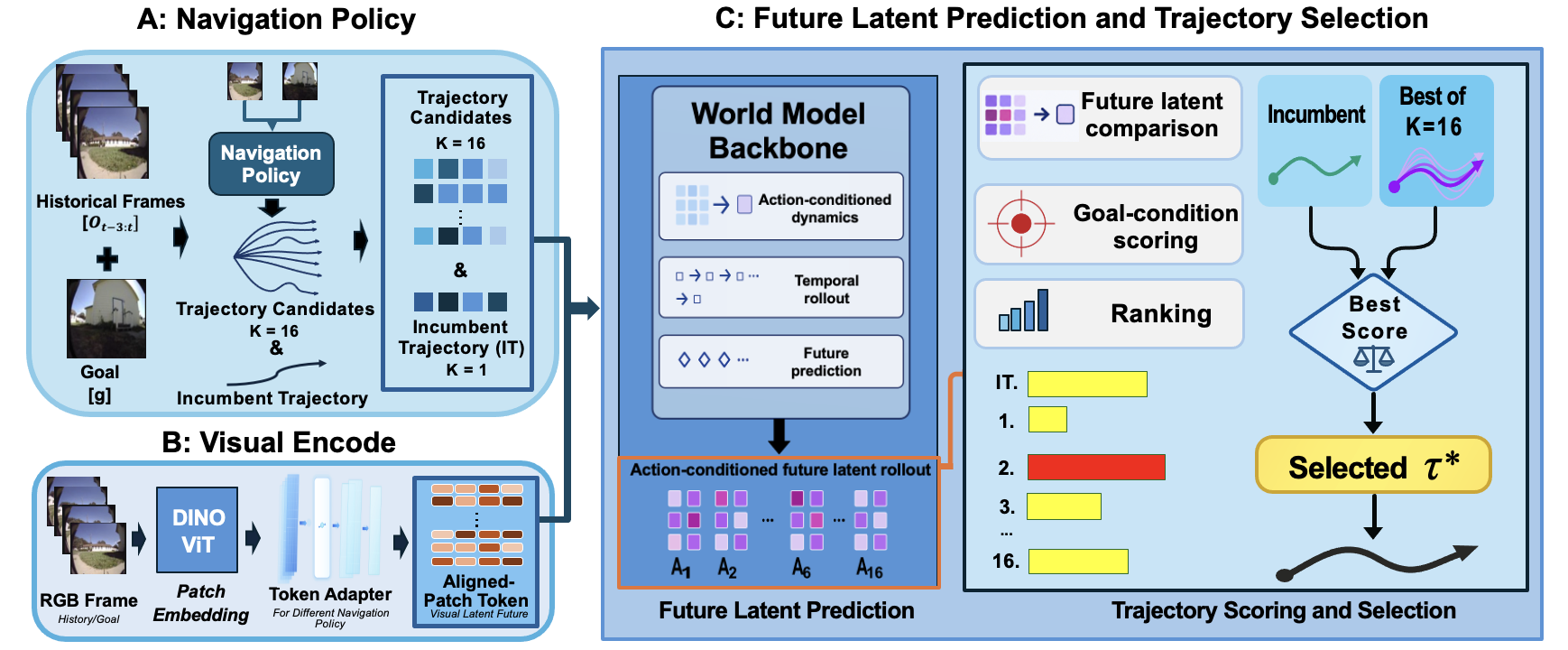}
    \caption{Overall workflow of \sys. \textbf{(A)} Given a four-frame
    observation history and a visual goal, a frozen goal-conditioned policy
    produces $K=16$ four-step trajectory challengers together with a retained
    incumbent $\tau_0$. \textbf{(B)} A shared frozen DINOv3 encoder and a
    lightweight adapter encode the history and goal once per state.
    \textbf{(C)} The action-conditioned world model jointly predicts H1--H4 future latents for the incumbent and all challengers. A goal-conditioned,
    proposal-aware scorer compares these predicted evolutions, ranks eligible
    challengers, and either selects the best one or retains the incumbent.
    Further details are provided in Sec.~\ref{subsec:selection}.}
    \label{fig:workflow}
\end{figure*}

\begin{figure}[!t]
    \centering
    \includegraphics[width=1\columnwidth]{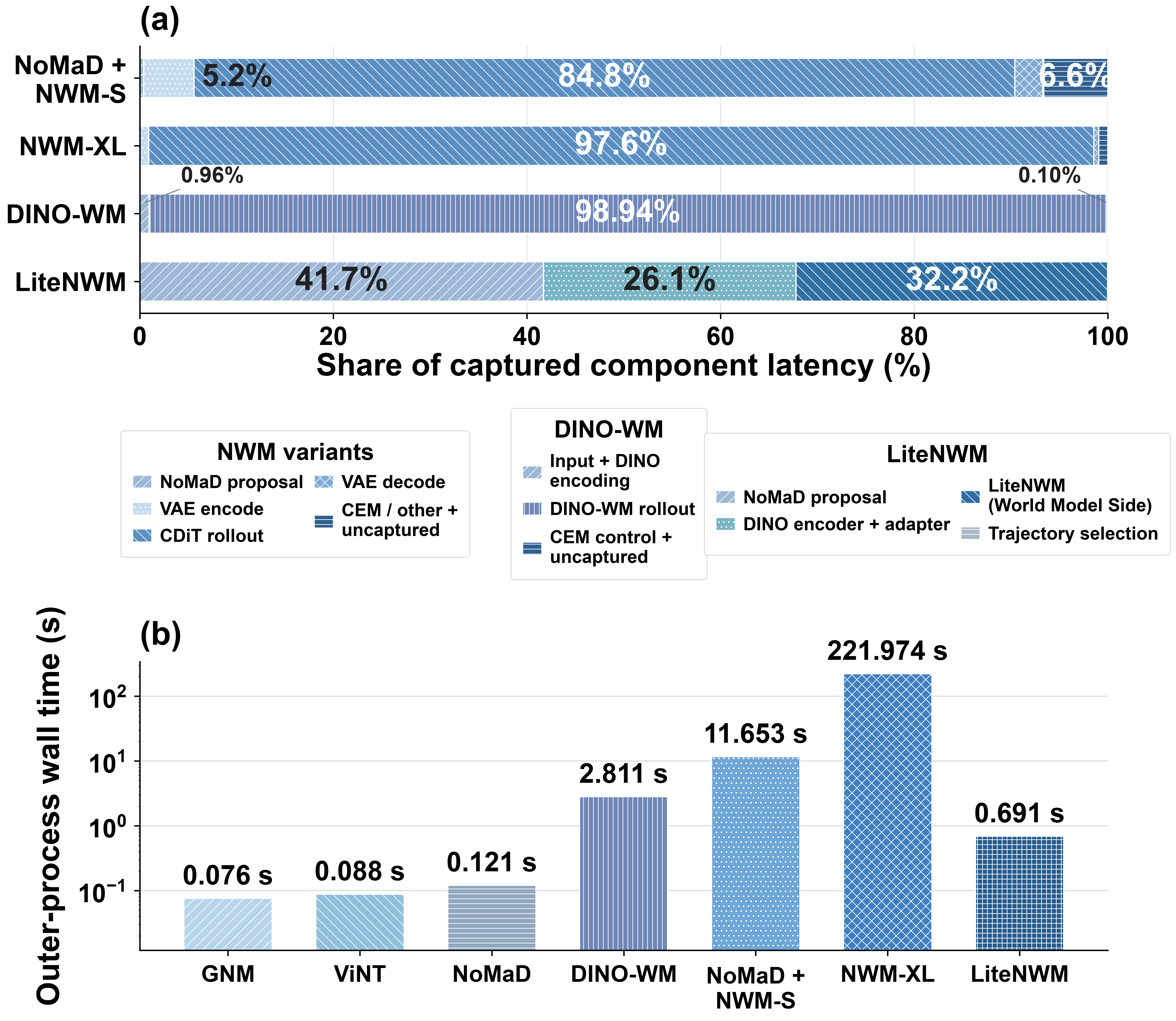}
    \caption{Captured component-latency profile under matched RTX~5090,
    RECON-100, $H=4$, and $K=16$ conditions. \textbf{(a)} Candidate-wise
    CDiT rollout accounts for $84.8\%$ and $97.6\%$ of the captured
    NoMaD+NWM-S and NoMaD+NWM-XL latency, respectively. \textbf{(b)} End-to-end outer-process wall time per state for the methods reported in Table III. LiteNWM amortizes candidate-wise RGB rollout through shared visual encoding, batched latent prediction, and trajectory scoring.}
    \label{fig:runtime_breakdown}
\end{figure}

\subsection{Latent Predictive and World--Action Models}
Latent world models provide an alternative by supporting decision making directly through learned predictive representations.
Such explicit visual generation also introduces computation that is not directly required by the downstream ranking objective, which is well established by previous work: PlaNet performs online planning in learned latent dynamics \cite{planet}, while TD-MPC2 optimizes trajectories using a decoder-free implicit world model \cite{tdmpc2}.
\par
More recently, DINO-WM predicts action-conditioned DINOv2 patch features for test-time action optimization \cite{dinowm}, V-JEPA~2-AC enables model-predictive control through action-conditioned latent prediction \cite{vjepa,vjepa2}, and DINO-world extends feature-space prediction to a generalist video model that can be adapted to observation--action trajectories \cite{dinoworld}.
DF$^{3}$ further demonstrates decoder-free feature forecasting, though its current formulation is observation-only and cannot model the consequences of alternative actions \cite{df3}.
ReL-NWM avoids RGB reconstruction through latent-space planning~\cite{relnwm}, but retains sequential autoregressive rollouts.
\sys targets rollout latency through shared visual encoding and joint multi-horizon prediction, eliminating autoregressive feedback across prediction horizons.
\par
In contrast, WAM-Nav and NavWAM take a different direction by coupling visual foresight with action generation inside unified world--action policies \cite{wamnav,navwam}.
\sys addresses a narrower question: whether action-conditioned latent prediction provides sufficient predictive evidence to distinguish among trajectory proposals from a frozen navigation policy.

\section{Method}
\label{sec:method}
Figure~\ref{fig:workflow} summarizes \sys, which uses frozen navigation policies, shared visual encoding, an action-conditioned latent predictor, and a trajectory scorer. 
In part C, the predictor estimates candidate futures, while the scorer evaluates them relative to the incumbent and proposal set.

\subsection{Problem Formulation}
\label{subsec:problem}

We consider goal-conditioned visual navigation. At time $t$, the robot
observes a four-frame visual history
$O_t=\{o_{t-3},o_{t-2},o_{t-1},o_t\}$ and a visual goal $g$. A frozen
navigation policy $\pi$ generates $K=16$ candidate trajectories. After a
deterministic waypoint-to-action conversion, each candidate is represented as
\begin{equation}
    \tau_i=\{a_{i,1},\ldots,a_{i,H}\}, \qquad
    a_{i,h}=[\Delta x,\Delta y,\Delta\psi]_{i,h}.
    \label{eq:trajectory}
\end{equation}
We retain an incumbent $\tau_0$ and form the complete comparison set
\begin{equation}
    \mathcal T_t=\{\tau_0,\tau_1,\ldots,\tau_K\}.
    \label{eq:candidate_set}
\end{equation}
Thus, $K$ denotes the number of challengers and does not include the
incumbent.

The latent predictor maps the observation history and each candidate action sequence to future features:
\begin{equation}
    \widehat Z_{i,1:H}=W_{\mathrm{latent}}(O_t,\tau_i).
    \label{eq:latent_world_model}
\end{equation}
The objective is to improve the policy's trajectory choice without replacing
or fine-tuning the proposer.


\subsection{Latent-space Navigation World Model}

The profile in Fig.~\ref{fig:runtime_breakdown} shows that candidate-wise RGB
rollout accounts for $84.8\%$ and $97.6\%$ of the captured NWM-S and NWM-XL latency, respectively. 
This motivates the evaluation path in Fig.~\ref{fig:workflow}(C): shared encoding and action-conditioned latent prediction replace candidate-wise RGB rollout before trajectory scoring.

\noindent\textbf{1) Trajectory Proposal and Shared Encoding.}
We use frozen NoMaD~\cite{nomad} only as the default navigation policy. A shared frozen
DINOv3 encoder~\cite{dinov3} and lightweight adapter encode the observation
history and goal once per state:
\begin{equation}
\begin{aligned}
    Z^{\mathrm{hist}} &= E_v(O_t), & Z^g &= E_v(g),\\
    (S_0,M_h) &= \mathcal H_{\rho}(Z^{\mathrm{hist}}).
\end{aligned}
\label{eq:visual_encoding}
\end{equation}
The history encoder $\mathcal H_{\rho}$ produces the last encoded state
$S_0$ and history memory $M_h$; $Z_t$ is the latest input token grid.
The history encoding is shared by all candidates. The navigation policy is conditioned on the visual goal, and the scorer uses its encoded feature; the latent predictor does not take the goal as input.

\begin{figure}[!t]
    \centering
    \includegraphics[width=0.98\columnwidth]{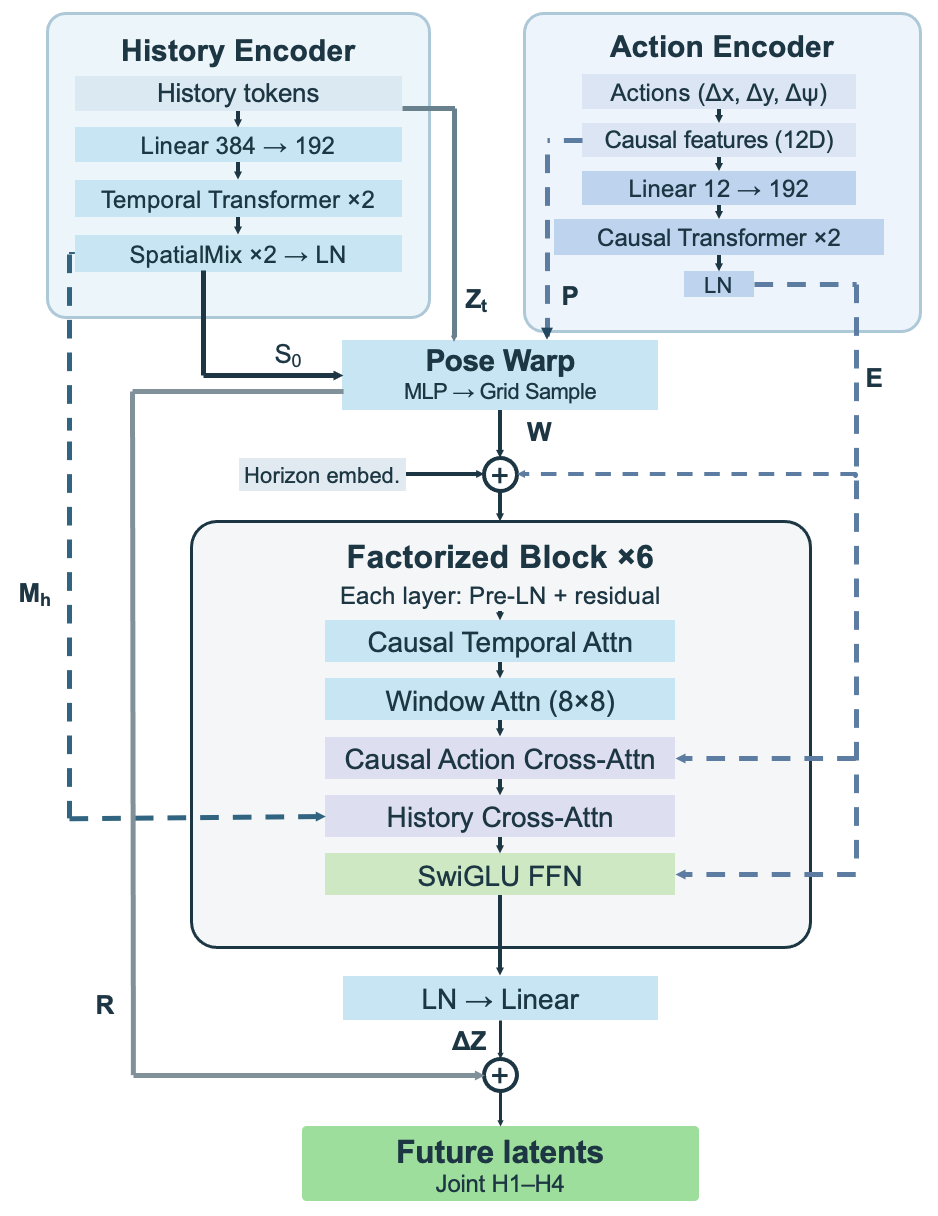}
    \caption{Backbone of \sys for joint H1--H4 latent prediction.
    Pose-conditioned warping provides $W$ for query construction and $R$
    for residual prediction. Six factorized blocks integrate action and
    history conditions. Dashed arrows denote conditioning.}
    \label{fig:LiteNav-WM}
\end{figure}

\noindent\textbf{2) Action-Conditioned Latent Prediction.}
Fig.~\ref{fig:LiteNav-WM} shows the backbone of \sys. For candidate $\tau_i$,
the action encoder returns action tokens and cumulative poses:
\begin{equation}
    (E_{i,1:H},P_{i,1:H})=A_{\eta}(\tau_i).
    \label{eq:action_encoding}
\end{equation}
The poses $P_{i,1:H}$ are computed by SE(2) integration. Pose-conditioned
warping initializes the queries:
\begin{equation}
\begin{aligned}
    W_{i,h} &= \operatorname{Warp}(S_0,P_{i,h}),\\
    R_{i,h} &= \operatorname{Warp}(Z_t,P_{i,h}),\\
    Q_{i,h}^{(0)} &= W_{i,h}+E_{i,h}+e_h,
\end{aligned}
\label{eq:query_construction}
\end{equation}
where $e_h$ is a learned horizon embedding; $E_{i,h}$ and $e_h$ are
broadcast across spatial tokens. Six factorized blocks refine the queries:
\begin{equation}
    Q_{i,1:H}^{(\ell)}=
    \mathcal B_{\ell}\!\left(
    Q_{i,1:H}^{(\ell-1)};E_{i,1:H},M_h
    \right), \quad \ell=1,\ldots,6 .
    \label{eq:rollout_blocks}
\end{equation}
Each sublayer uses Pre-LN and a learned residual scale. Temporal and
action attention are causal, and action tokens modulate SwiGLU.
An LN--linear head predicts the residual:
\begin{equation}
    \widehat Z_{i,h}=
    R_{i,h}+D_{\delta}\!\left(Q_{i,h}^{(6)}\right),
    \qquad h=1,\ldots,H .
    \label{eq:LiteNav-WM_forward}
\end{equation}
All four horizons are predicted jointly without autoregressive feedback.
The latent predictor is supervised by frozen DINOv3-adapted future features.
\label{subsec:litewm}
\noindent\textbf{3) Future-Aware Trajectory Scoring.}
The scorer combines predicted futures with the current state and goal:
\begin{equation}
    \phi_i=\phi(\widehat Z_{i,1:H},Z_t,Z^g).
    \label{eq:future_features}
\end{equation}
Together with action, history, and proposal-relative features,
it predicts outcome changes from the incumbent:
\begin{equation}
    \Delta y_i=y_i-y_0=
    [\Delta\mathrm{ATE},\Delta\mathrm{RPE}_{\mathrm{trans}},
     \Delta\mathrm{FDE},\Delta J]_i.
    \label{eq:relative_targets}
\end{equation}
RPE denotes one-step translational RPE;
$\mathrm{FDE}_i=\|p_{i,H}-p_H^{\mathrm{ref}}\|_2$ and
$J_i=0.4\,\mathrm{ATE}_i+0.6\,\mathrm{FDE}_i$.
Targets use metres, whereas reported errors are normalized by
dataset waypoint spacing. Negative changes indicate improvement.
Reference future poses are unavailable during selection.

\subsection{Training, Trajectory Selection, and Execution}
\label{subsec:selection}

\noindent\textbf{Training and calibration.}
We sequentially train three-member predictor and scorer ensembles,
keeping NoMaD and DINOv3 frozen.
Predictor losses comprise latent regression, cosine,
action-sensitivity, temporal-consistency, and uncertainty terms.
Scorer losses comprise heteroscedastic regression,
dual-improvement/harm classification, and pairwise/listwise ranking.
Both stages use AdamW with weight decay $10^{-4}$.
The predictor uses learning rate $6\times10^{-5}$, batch size 56,
gradient clipping at 1.0, and 300 warm-up steps followed by cosine decay.
The scorer uses learning rate $10^{-4}$, clipping at 5.0,
and early stopping on the disjoint scorer-calibration partition.

\noindent\textbf{Refinement and selection.}
CEM1 performs one refinement iteration~\cite{cem}, with all models frozen.
For $M=3$ ensemble members,
\begin{equation}
\begin{aligned}
    \bar\mu_{i,q}
    &=\frac{1}{M}\sum_m\mu_{i,q}^{(m)},\\
    \sigma_{i,q}^{2}
    &=\frac{1}{M}\sum_m
    \left[(s_{i,q}^{(m)})^2+
    (\mu_{i,q}^{(m)}-\bar\mu_{i,q})^2\right].
\end{aligned}
\label{eq:total_variance}
\end{equation}
The conservative estimate is
\begin{equation}
    u_{i,q}=\bar\mu_{i,q}+z_d\sigma_{i,q},
    \label{eq:calibrated_outcome}
\end{equation}
where $z_d$ is shared across outcomes within dataset $d$.
For $k\in\{\mathrm{dual},\mathrm{harm}\}$, calibrated probabilities
are $p_i^k=\operatorname{sigmoid}(\ell_i^k/T_d^k)$,
using ensemble-averaged logits.
Dual improvement requires decreases in both ATE and $J$;
harm denotes an increase in either, with tolerance $10^{-7}$ m.

Using ranking logits $r_i$, define
\[
\widetilde r_i=
\frac{r_i-r_0}{\max\{\operatorname{Std}_{j=1}^{K}(r_j),10^{-3}\}},
\]
where $\operatorname{Std}$ uses population normalization.
The selection score is
\[
\begin{aligned}
e_i={}&u_{i,\mathrm{ATE}}
 +0.25u_{i,\mathrm{RPE}_{\mathrm{trans}}}
 +0.25[u_{i,\mathrm{FDE}}]_+\\
 &+0.50u_{i,J}+0.15p_i^{\mathrm{harm}}
 -0.05p_i^{\mathrm{dual}}-0.03\widetilde r_i,
\end{aligned}
\]
with $[x]_+=\max(x,0)$.
A challenger is eligible if it passes physical and proposal-support
checks and satisfies
\[
\begin{aligned}
u_{i,\mathrm{ATE}}&<-m_d^{\mathrm{ATE}},
&u_{i,J}&<-m_d^J,\\
p_i^{\mathrm{dual}}&\geq\theta_d^{\mathrm{dual}},
&p_i^{\mathrm{harm}}&\leq\theta_d^{\mathrm{harm}}.
\end{aligned}
\]
This eligibility test defines the learned acceptance gate.
Temperatures, $z_d$, and thresholds are fitted on scorer-calibration data.
Let $\mathcal E_t$ contain the eligible challengers. We select
\begin{equation}
    \widehat i=
    \begin{cases}
        \displaystyle\arg\min_{i\in\mathcal E_t}e_i,
        &\mathcal E_t\neq\emptyset,\\
        0,&\mathcal E_t=\emptyset,
    \end{cases}
    \qquad\tau^*=\tau_{\widehat i}.
    \label{eq:selective_replacement}
\end{equation}
Otherwise, the incumbent is retained.

\noindent\textbf{Receding-horizon execution.}
At each control cycle, \sys observes, proposes, predicts, selects,
executes a short segment of $\tau^*$, and replans. For a different navigation policy
such as MBRA~\cite{mbra}, a deterministic action adapter maps its output to
the same four-step local representation in Eq.~\eqref{eq:trajectory}. All
\sys components remain frozen during deployment.

\raggedbottom
\setlength{\textfloatsep}{8pt plus 2pt minus 2pt}
\setlength{\dbltextfloatsep}{8pt plus 2pt minus 2pt}
\setlength{\floatsep}{6pt plus 2pt minus 2pt}
\setlength{\dblfloatsep}{6pt plus 2pt minus 2pt}
\makeatletter
\setlength{\@fptop}{0pt}
\setlength{\@dblfptop}{0pt}
\makeatother

\clearpage

\begin{strip}
    \MainResultsTable
\end{strip}

\section{Experiments}
\label{sec:experiments}


We evaluate trajectory selection, predictive representations, end-to-end efficiency, and physical-robot deployment.


\begin{table*}[!t]
    \AuxResultsTables
\end{table*}

\subsection{Experimental Setup}
\label{subsec:experimental_setup}

\paragraph{Datasets and protocol}
We use fixed development cohorts from RECON~\cite{recon}, SCAND~\cite{scand}, and SACSoN~\cite{sacson}.
RECON contains diverse open-world navigation trajectories, SCAND spans indoor and outdoor social navigation on different robot platforms, and SACSoN emphasizes indoor robot--pedestrian interaction.
Each domain contains 100 query states with a four-step prediction horizon. 
Four proposal seeds $\{23,42,73,101\}$ produce 1,200 state--seed evaluations. 
\sys starts from the same frozen pool of $K=16$ NoMaD challengers, while the deterministic NoMaD output is retained as the incumbent. 

\paragraph{Baselines}
We select different visual navigation baselines and their combination to showcase the effectiveness of \sys:
\begin{itemize}
\setlength{\itemsep}{0pt}
\setlength{\parsep}{0pt}
\item \textbf{Navigation policy baselines:}
GNM~\cite{gnm}, ViNT~\cite{vint}, and NoMaD~\cite{nomad}, which produce
navigation actions without explicit future-aware reranking.
\item \textbf{Generative navigation world models:}
NWM-S, NWM-XL, NoMaD+NWM-S and NoMaD+NWM-XL~\cite{nwm}, which evaluate the frozen
NoMaD candidates through generated future observations.
\item \textbf{Latent world model baseline:}
DINO-WM~\cite{dinowm} is adapted to rerank trajectory candidates using predicted future DINO features.
\end{itemize}

\paragraph{Metrics and statistics}
Trajectories are expressed in the query-centered robot frame and normalized by dataset-average waypoint spacing (RECON:
$0.25\,\mathrm{m}$; SCAND: $0.38\,\mathrm{m}$; SACSoN:
$0.255\,\mathrm{m}$). Lower ATE and RPE indicate better trajectory accuracy.

\begin{itemize}
\setlength{\itemsep}{0pt}
\setlength{\parsep}{0pt}
\setlength{\topsep}{2pt}

\item \textbf{ATE:}
Translational RMSE over the $H$ future waypoints.

\item \textbf{RPE:}
Relative-translation RMSE between successive waypoints.

\item \textbf{SR@1:}
Percentage of trajectories with normalized $\mathrm{FDE}\leq1$;
higher is better. It measures offline endpoint agreement rather than
closed-loop navigation success.
\end{itemize}

\begin{figure}[!t]
    \centering
    \includegraphics[width=\columnwidth]{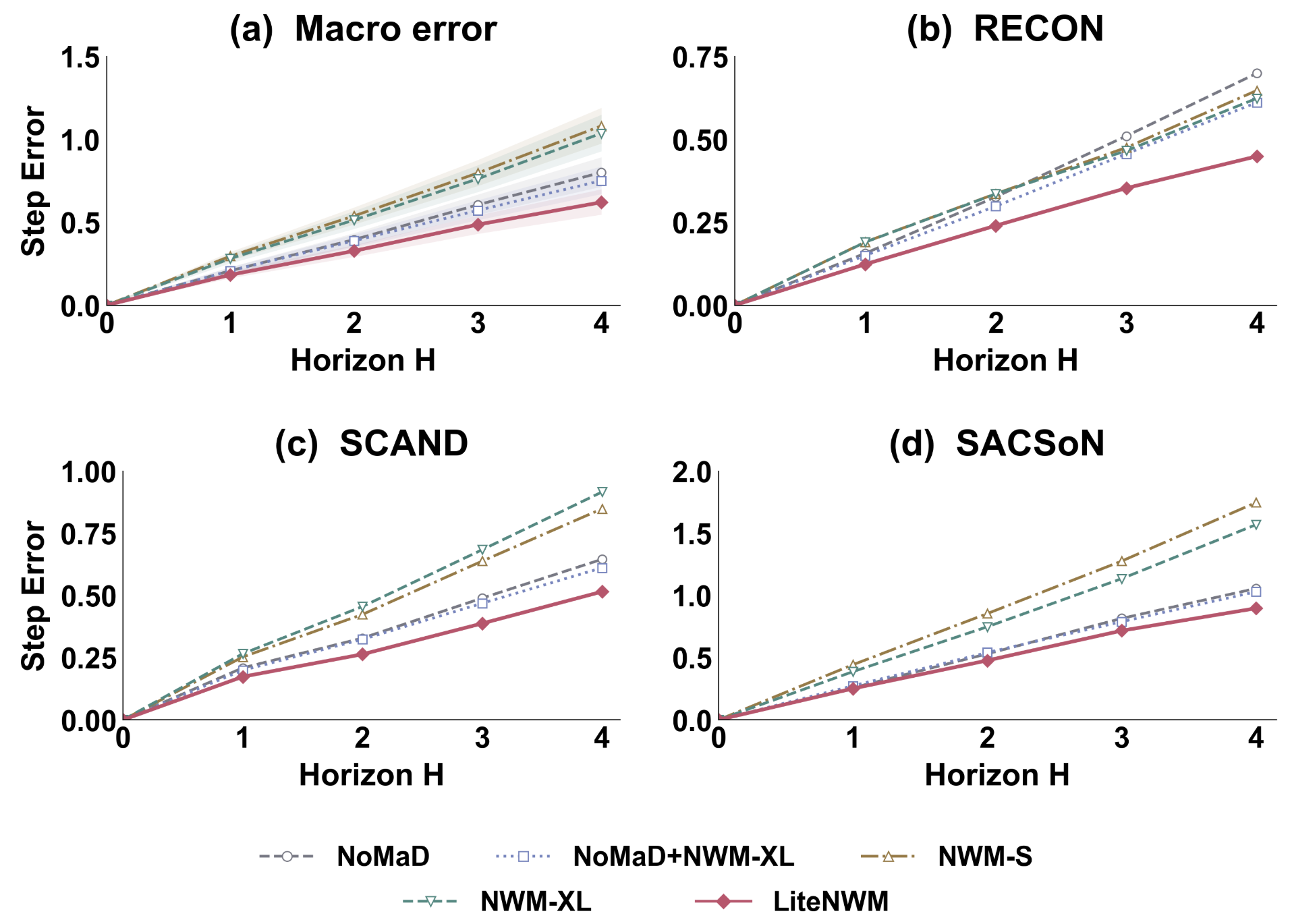}
    \caption{Step error over the prediction horizon (H). The curves report the
    macro error and the errors on RECON, SCAND, and SACSoN for NoMaD,
    NoMaD+NWM-XL, and \sys. LiteNWM shows a consistently slower error
    increase at later timesteps across all three domains.}
    \label{fig:step_error}
\end{figure}

No post-hoc alignment or path-length normalization is applied.
Absolute errors are not directly comparable with the original NWM results,
which use length-eight trajectories~\cite{nwm}, whereas our matched evaluation
uses $H=4$. 
NWM’s MPC/CEM and terminal-frame objective were designed for eight-step rollouts, which may limit their effectiveness when adapted to $H=4$ short-horizon trajectories~\cite{nwm}.
Optimizing terminal-frame visual similarity may overlook intermediate trajectory deviations, potentially contributing to NWM's higher ATE.
As shown in Fig.~\ref{fig:step_error}, per-step error and trajectory drift accumulate with prediction time, so the four additional
late-horizon waypoints at $H=8$ produce substantially larger
trajectory-level errors than at $H=4$.

\subsection{End-to-end Evaluation}
\label{subsec:cross_domain}

Table~\ref{tab:main_results} provides the main system-level comparison.
Relative to NoMaD+NWM-XL, \sys reduces ATE by 25.21\% on RECON, 14.98\% on SCAND, and 14.71\% on SACSoN.
\sys obtains the lowest ATE and RPE and the highest SR@1 in all three domains. Compared with NoMaD+NWM-XL, it raises SR@1 by 6.5, 3.5, and 7.0 percentage points on RECON, SCAND, and SACSoN, respectively.

Figure~\ref{fig:step_error} shows that \sys reduces ATE by slowing error
growth over the prediction horizon. Although the methods have similar errors
at the initial timestep, the baseline errors increase more rapidly at later
timesteps across all three domains. Because ATE aggregates errors over four
future waypoints, this reduced accumulation leads to lower ATE
(Table~\ref{tab:main_results}). NWM performance may also be affected by the
$H=8$-to-$H=4$ horizon and prior mismatch, the imperfect alignment between terminal
LPIPS and geometric accuracy, and the official one-step CEM update. In
contrast, joint H1--H4 prediction provides intermediate and terminal evidence
for rejecting candidates with accumulating errors. The next section evaluates the contribution of joint H1--H4 prediction to
trajectory selection.

\subsection{Ablation Study}
\label{subsec:ablation}

Table~\ref{tab:ablation} compares four representations on identical
candidate pools. All variants retain the same observed visual context
and candidate actions, with scorers trained separately under matched
conditions.
\emph{No predicted future} removes the future-feature block;
\emph{Static DINO repeat} replaces it with repeated current DINO features;
\emph{Predicted H4 only} uses only terminal-prediction features; and
\emph{Predicted H1--H4} uses all four predicted horizons.
Predicted futures improve selection over both controls.
H4 captures most of this gain; intermediate horizons provide
a smaller additional improvement by supplying trajectory-evolution
features beyond the terminal state.

To test whether the scorer uses the future associated with each
candidate, we freeze the H1--H4 scorer and apply two prespecified permutations of candidate--future pairings.
Complete predicted-future feature blocks are permuted while actions,
context, and model weights remain fixed; derived features are recomputed.
Averaged over the two permutations, macro ATE increases by $0.0294$
(95\% paired cluster-bootstrap CI: $[0.0175,\,0.0415]$).
The interval uses 20,000 dataset-stratified cluster resamples,
retaining grouped states and proposal seeds.
As consistency checks, jointly permuting complete candidate bundles
preserves selections, and the no-prediction control is also unaffected.
These results support the scorer's use of correctly paired
predicted-future evidence.

\begin{figure*}[!t]
    \centering
    \includegraphics[width=0.90\textwidth]{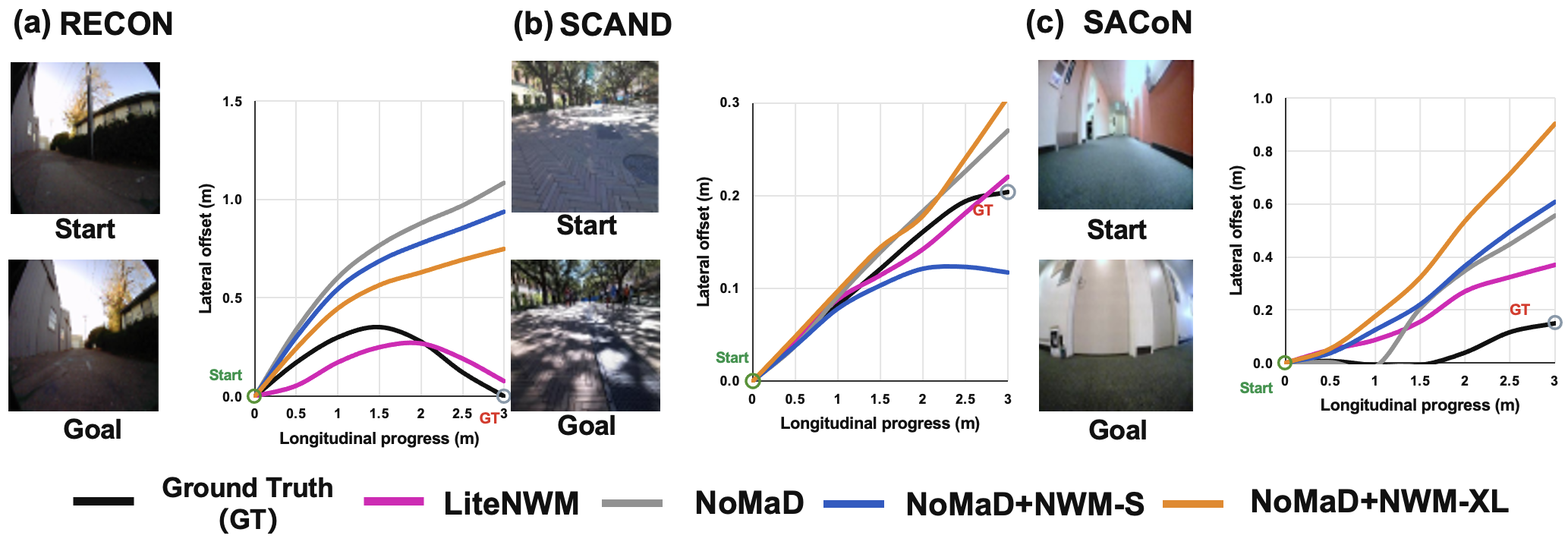}
    \caption{Representative offline trajectory selections on three datasets. For each dataset, the left panels show the start and goal
    observations for NoMaD+NWM-S, NoMaD+NWM-XL, and \sys, while the
    right panel compares the selected trajectories with ground truth over the
    first 3\,m of longitudinal progress.}
    \label{fig:offline_examples}
    \label{fig:offline_cases}

    \vspace{0.35\baselineskip}
    \includegraphics[width=0.95\textwidth]{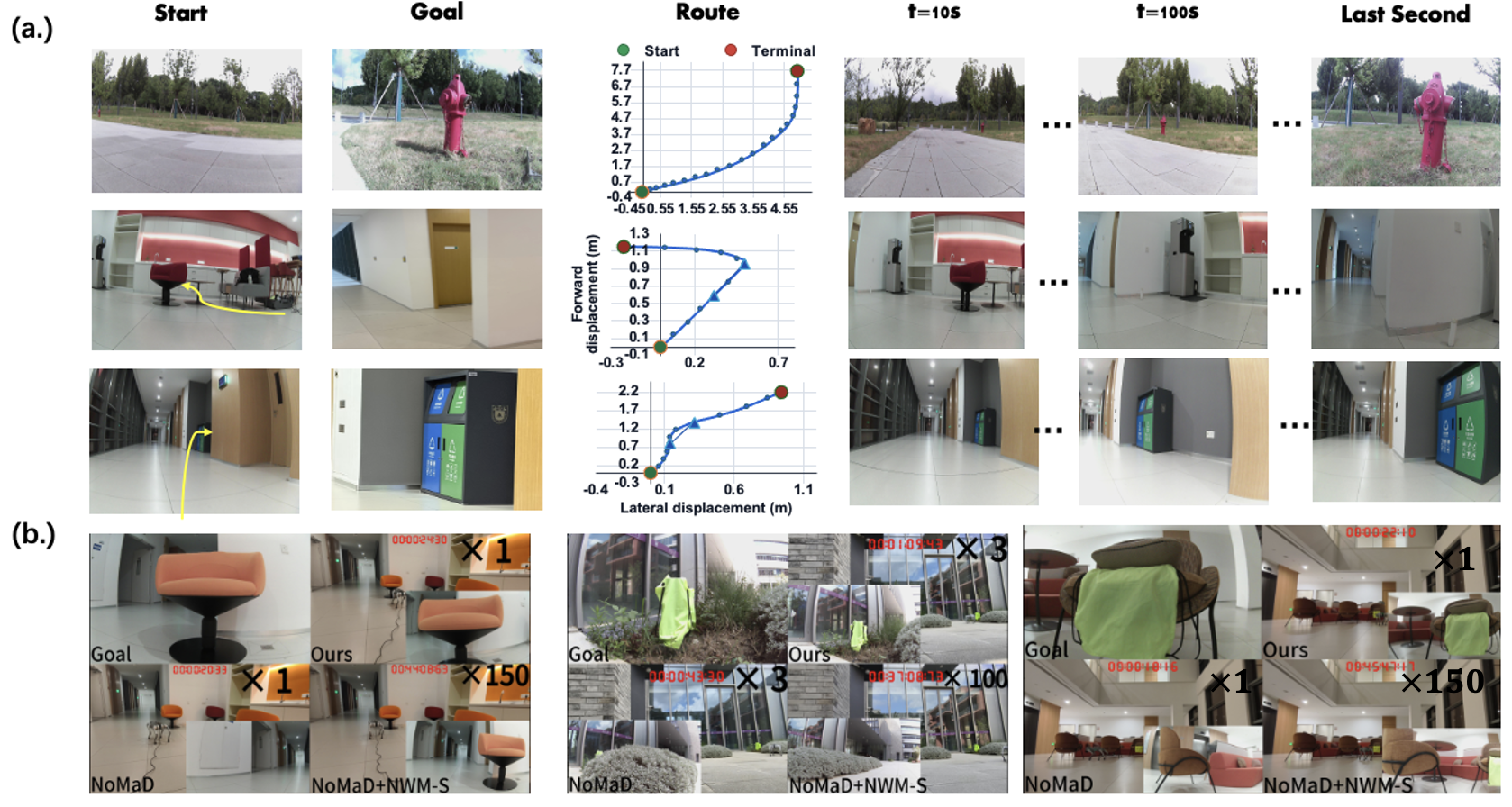}
    \caption{(a) Indoor and outdoor routes with start, intermediate, and goal observations. (b) Individual comparison runs, including exploratory NoMaD+NWM-S deployment probes.}
    \label{fig:real_world}
    \label{fig:real_world_rollouts}
\end{figure*}

\subsection{Transfer Across Navigation Policies}
\label{subsec:cross_proposer}

Table~\ref{tab:mbra_transfer} replaces NoMaD with MBRA while keeping the \sys frozen. 
A deterministic adapter maps MBRA trajectories to the four-step local action representation used by \sys, without proposer-specific training or recalibration.

The frozen evaluator reduces MBRA ATE by 6.9\% on RECON, 23.0\% on SCAND, and 16.8\% on SACSoN.
The adapter only standardizes the action interface; candidate selection is still performed by the original \sys and scorer.

This transfer follows from the navigation policy--evaluator separation in Sec.~\ref{sec:method}. 
The result supports transfer across the different navigation policies.

\begin{center}
\begin{minipage}{\columnwidth}

\refstepcounter{table}
\label{tab:mbra_transfer}

{\footnotesize
\noindent
\parbox{\linewidth}{%
\raggedright
\textbf{TABLE~\thetable}\enspace
Transfer of \sys to another navigation policy, MBRA.
Macro is the three-domain mean. Lower is better.
\par
}}

\vspace{0.45em}

\centering
    \setlength{\tabcolsep}{3.5pt}
  \scriptsize
  \DWMUniformTable{\columnwidth}{\scriptsize}{3.5pt}{5}{l|cccc}{
        \hline
        Method & RECON & SCAND & SACSoN & Macro \\
        \hline
        MBRA & 0.408 & 0.482 & 0.911 & 0.600 \\
        \textbf{MBRA+\sys}
        & \textbf{0.380} & \textbf{0.371}
        & \textbf{0.758} & \textbf{0.503} \\
        \hline
        Relative reduction & 6.9\% & 23.0\% & 16.8\% & 16.2\% \\
        \hline
    }
  \end{minipage}
\end{center}

\subsection{System Performance}
\label{subsec:efficiency}

We measure end-to-end inference on an NVIDIA RTX~5090 using the same hardware, cohort, and timing boundary for all methods.
The timing boundary includes proposal generation, future evaluation, scoring, refinement, and final selection.

As shown in Table~\ref{tab:efficiency}, \sys is 16.86$\times$ faster than NoMaD+NWM-S and 128.00$\times$ faster than NoMaD+NWM-XL. 
A separate seven-state native-CEM diagnostic measures standalone NWM-XL at 221.974~s per state, or 321.24$\times$ the \sys latency; no energy value was collected for this diagnostic.
Figure~\ref{fig:runtime_breakdown} localizes the difference to the candidate-conditioned image-rollout stage, which dominates the NWM runtime.

The profile in Fig.~\ref{fig:runtime_breakdown} links this speedup to the evaluation path: candidate-wise RGB rollout accounts for 84.8\% and 97.6\% of the captured NWM-S and NWM-XL component latency. \sys encodes the history and goal once per state, jointly predicts candidate futures across H1--H4 without autoregressive feedback, and scores these futures without RGB decoding. This replaces the dominant image-rollout stage with shared encoding, joint prediction, and relative evaluation. The end-to-end measurements in Table~\ref{tab:efficiency} include proposal generation, refinement, and final selection; the gain therefore reflects the complete pipeline rather than predictor latency alone.

\subsection{Real-World Deployment}
\label{subsec:real_world}

We deploy \sys on a Unitree Go2 EDU.
The navigation policies receive only monocular RGB observations
and an RGB goal image.
We evaluate 10 trials per method in each of the three scenes
listed in Table~\ref{tab:real_world_results}.
Success requires entering a calibrated 1.0-m goal radius with
a visible, goal-consistent target observation and yaw error
within $60^\circ$, before collision, corrective intervention,
or abnormal termination.
The valid state must persist for at least 1~s or be followed
by a normal stop; merely passing the target does not count.

\noindent\textbf{Transfer to unseen layouts.}
\sys performs closed-loop navigation in previously unseen
indoor and outdoor layouts without scene-specific training
or online model updates.
Figure~\ref{fig:real_world}(a) additionally illustrates
representative trajectories through furniture-constrained
spaces, outdoor areas, and corridors.
These examples provide qualitative evidence of deployment
across changes in scene structure, illumination, and viewpoint.

\noindent\textbf{Reliable closed-loop goal reaching.}
\sys achieves more reliable goal reaching than NoMaD across the three evaluated settings
(Table~\ref{tab:real_world_results}).
Figure~\ref{fig:real_world}(b) shows representative comparisons in the open indoor, cluttered indoor, and outdoor courtyard scenes.
The larger success-rate gains in the cluttered indoor route and outdoor courtyard suggest that the benefit extends beyond the short open-indoor route.
This pattern is consistent with the future-aware selection design in Sec.~\ref{sec:method}.
\sys continuously refreshes the visual history and predicts the H1--H4 latent consequences of proposed trajectories.
The scorer relates these predicted futures to the visual goal and retained incumbent, providing evidence about intermediate motion and terminal goal consistency before execution.
This receding-horizon procedure repeatedly incorporates the latest observation, allowing future-aware comparisons to be updated as the robot moves.


\noindent\textbf{Efficient future-aware navigation.}
\sys reduces the computational burden of future-aware selection, as quantified by the profiling in Table~\ref{tab:efficiency}.
The deployment probes reveal the practical scale of the
NWM-S planning delay: even with DDIM reduced to 30 steps,
the two runs average 234.0 and 236.5~s per planning decision,
over 13 and 9 decisions, respectively.
These exploratory runs characterize computational cost and
are excluded from aggregate navigation-outcome comparisons.
The lower evaluation cost follows from the computational
design in Sec.~\ref{sec:method}: the history and goal are
encoded once and shared across candidates, while candidate
futures are evaluated jointly in latent space without RGB
decoding.
This preserves candidate-conditioned look-ahead while avoiding
the expensive image-generation stage at each replanning step.
\begin{center}
  \begin{minipage}{\columnwidth}
\refstepcounter{table}
\label{tab:real_world_results}

{\footnotesize
\noindent
\parbox{\linewidth}{%
\raggedright
\textbf{TABLE~\thetable}\enspace
Real-world image-goal navigation.
Each method has 10 trials per scene and 30 trials overall.
SR denotes episode success rate; SPL averages all trials,
assigning zero to failures.
\par
}}

\vspace{0.45em}

\centering
    \begingroup
\fontsize{7.2}{8.6}\selectfont
\renewcommand{\arraystretch}{1.05}
\setlength{\tabcolsep}{6pt}
\setlength{\heavyrulewidth}{0.5pt}
\setlength{\lightrulewidth}{0.4pt}
\setlength{\cmidrulewidth}{0.3pt}
\begin{tabular}{@{}lcccc@{}}
  \toprule
  \multirow{2}{*}{Scene}
  & \multicolumn{2}{c}{NoMaD}
  & \multicolumn{2}{c}{\sys} \\
  \cmidrule(lr){2-3}\cmidrule(lr){4-5}
  & SR (\%) & SPL & SR (\%) & SPL \\
  \midrule
  Open indoor
  & 70.0 & 0.642 & \textbf{90.0} & \textbf{0.890} \\
  Cluttered indoor
  & 40.0 & 0.397 & \textbf{90.0} & \textbf{0.889} \\
  Outdoor courtyard
  & 20.0 & 0.187 & \textbf{70.0} & \textbf{0.651} \\
  \midrule
  Overall
  & 43.3 & 0.409 & \textbf{83.3} & \textbf{0.810} \\
  \bottomrule
\end{tabular}
\endgroup
  \end{minipage}
\end{center}
\section{Conclusion}
Combining action-conditioned multi-horizon latent prediction with incumbent-relative scoring, \sys selects trajectories from a frozen policy without RGB reconstruction.
On matched initial NoMaD proposal pools across RECON, SCAND, and SACSoN, it reduces macro ATE by 17.56\% over NWM-XL.
Ablations support the contribution of predicted future features beyond static visual context.
Experiments on RTX~5090 show 18.39$\times$ and 132.02$\times$ end-to-end speedups over NWM-S and NWM-XL, respectively.
The frozen evaluator transfers to other navigation policies, reducing MBRA macro ATE by 16.2\% without proposer-specific retraining.
In real-world experiments across three scenes, each method is evaluated over 30 trials on a Unitree Go2 EDU.
\sys improves the navigation success rate from 43.3\% to 83.3\% over NoMaD and increases SPL from 0.409 to 0.810.
Future work will extend action-conditioned latent foresight beyond navigation to manipulation tasks.


\providecommand{\LiteAppendixTableCaption}[2]{%
  \refstepcounter{table}\label{#1}%
  {\footnotesize\noindent
  \parbox{\linewidth}{\raggedright
    \textbf{TABLE~\thetable}\enspace #2\par}}%
  \par\vspace{0.45em}%
}

\graphicspath{{./}{figures/}}
\appendices

\section{Supplementary Evaluation Protocol}
\label{app:protocol}

The supplementary analyses follow the query, proposal-seed, and metric
definitions in Sec.~\ref{subsec:experimental_setup}. Macro results give
equal weight to the three domains. The fixed-pool representation and
inference-intervention diagnostics disable CEM and the learned acceptance
gate while retaining action-validity checks. Representation variants use
separately trained, matched scorers while the inference interventions keep the
diagnostic H1--H4 model fixed. Separately from the main-text
ablation, supplementary representation contrasts are
computed within their own matched runs. The fixed-pool search control in Appendix~\ref{app:search}
retains the gate.

For the four-seed experiments, paired bootstrap resampling keeps all
queries and proposal seeds within each sampled group. Confidence
intervals (CIs) are 95\% cluster-bootstrap intervals, and Holm corrections
use the original experiment families. Historical evaluations retain
their original configurations and candidate pools, as specified below.
RPE is computed from the three consecutive differences between the four
future poses, without adding the current-pose origin; SR@$r$ denotes
the percentage of normalized FDE values at most $r$.

\section{Predictive Features and Action Conditioning}
\label{app:mechanism}

\subsection{Predictor Action Input}

To isolate predictor-side action conditioning, we permute challenger
actions at the predictor while keeping model weights, visual inputs,
the incumbent, actual candidate trajectories, and scorer-side action
descriptors fixed. 
Permutation increases macro ATE by 0.0300 and changes 63.75\% of selections. The error increase shows that predictor-side actions contribute to candidate evaluation even when the scorer's action descriptors are unchanged.

Permutation changes 70.00\%, 42.50\%, and 78.75\% of selections on RECON, SCAND, and SACSoN, respectively. Conditional on a changed selection, mean ATE increases by 0.0549, 0.0156, and 0.0569. 
SCAND has both the lowest change rate and the smallest conditional increase, consistent with its smaller aggregate effect of 0.0066 with comparing by 0.0385 on RECON and 0.0448 on SACSoN.

\begin{figure}[!t]
\centering
\includegraphics[width=\columnwidth]{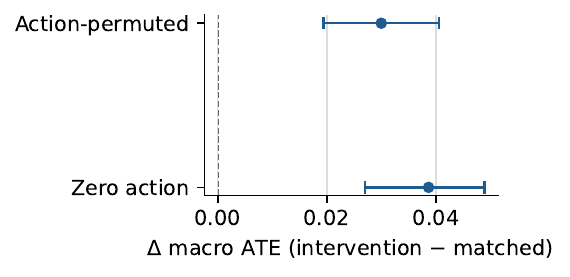}
\caption{Predictor-action interventions in the fixed-pool diagnostic.
Points show intervention minus matched-action macro ATE; bars show
95\% paired cluster-bootstrap CIs using 100/7/100 groups over
300 queries and four proposal seeds.
Model weights, visual inputs, candidate trajectories, and scorer-side
action descriptors remain fixed.}
\label{fig:app_action}
\end{figure}

Figure~\ref{fig:app_action} also reports a zero-action intervention, which increases macro ATE by 0.0386.
Permutation preserves the sampled action values while changing their assignment.
Zeroing additionally changes the input distribution makes it a secondary sensitivity check.

\subsection{Selection Changes and Dependence Grouping}

The candidate--future correspondence intervention described in
Sec.~\ref{subsec:ablation} changes 62.46\% of selections.

H1--H4 and H4-only select the same candidate in 65.08\% of
state--seed pairs. The H4-only minus H1--H4 ATE contrast is
$+0.0031$, with a CI spanning zero
(Table~\ref{tab:app_diagnostics}).

Table~\ref{tab:app_diagnostics} tests sensitivity to the dependence grouping used for uncertainty estimation. 
Resampling preserves all queries and four proposal seeds within each group, with query weighting within domains and equal domain weighting for macro estimates. 
The intervals remain positive for removing predicted futures, mismatching future features, and permuting predictor actions.
Therefore, the evidence for predictive-feature persists under this alternative grouping of the same cohorts.

\begin{table}[!t]
\centering
\LiteAppendixTableCaption{tab:app_diagnostics}{
Supplementary diagnostic contrasts under alternative grouping.
$\Delta$ATE is reference or intervention minus the matched
H1--H4 baseline; positive values favor
the baseline. CIs use 20,000 paired cluster-bootstrap resamples
and are retrospective sensitivity intervals.}
\DWMUniformTable{\linewidth}{\scriptsize}{3pt}{3}{l|cc}{
  \hline
  Reference or intervention & $\Delta$ATE & 95\% CI \\
  \hline
  No predicted future & $+0.0158$ & $[0.0052,0.0256]$ \\
  H4-only             & $+0.0031$ & $[-0.0014,0.0065]$ \\
  Future mismatch     & $+0.0294$ & $[0.0175,0.0413]$ \\
  Action permutation  & $+0.0300$ & $[0.0169,0.0424]$ \\
  \hline
}
\end{table}

\section{Search Budget and Selection Quality}
\label{app:search}

\subsection{Selection Within the Initial Pool}

Fixed-pool selection retains most of the default configuration's
observed error reduction relative to NoMaD+NWM-XL. Using the macro values in
Table~\ref{tab:app_search}, fixed-pool selection retains 94.75\%,
95.03\%, and 93.32\% of the CEM1 error reductions relative to
NoMaD+NWM-XL for ATE, RPE, and FDE, respectively. Each percentage is
the baseline-to-fixed-pool error difference divided by the
baseline-to-CEM1 difference.

The fixed-pool selection diagnostic also compares selected trajectories with the retained incumbent.
A challenger with lower ATE than the incumbent exists in 98.42\% of the 1,200 state--seed observations. 
The selected output improves ATE in 49.75\%, increases it in 8.42\%, and retains the incumbent in 41.83\%.
Among the 698 replacements, 597 (85.53\%) contribute to the improvement of trajectory accuarcy.

Within the initial pool's top-1, top-3, and top-5 candidates, ground-truth ATE ranks place 37.25\%, 68.05\%, and 80.66\% of replacements respectively.
\begin{table}[!t]
\centering
\LiteAppendixTableCaption{tab:app_search}{
Three-domain macro results by refinement budget.
Methods share the initial challenger pool; subsequent CEM budgets differ.
Lower errors are better. Bold marks the default configuration.}
\DWMUniformTable{\linewidth}{\small}{5pt}{4}{l|ccc}{
  \hline
  Configuration & ATE & RPE & FDE \\
  \hline
  NoMaD+NWM-S & 0.5489 & 0.2805 & 0.7781 \\
  NoMaD+NWM-XL & 0.5538 & 0.2813 & 0.7794 \\
  \hline
  LiteNWM, fixed pool & 0.4617 & 0.2488 & 0.6453 \\
  \textbf{LiteNWM, CEM1}
    & \textbf{0.4566} & \textbf{0.2471} & \textbf{0.6357} \\
  LiteNWM, CEM3 & 0.4537 & 0.2469 & 0.6290 \\
  \hline
}
\end{table}

\begin{figure*}[!t]
\centering
\includegraphics[width=\textwidth]{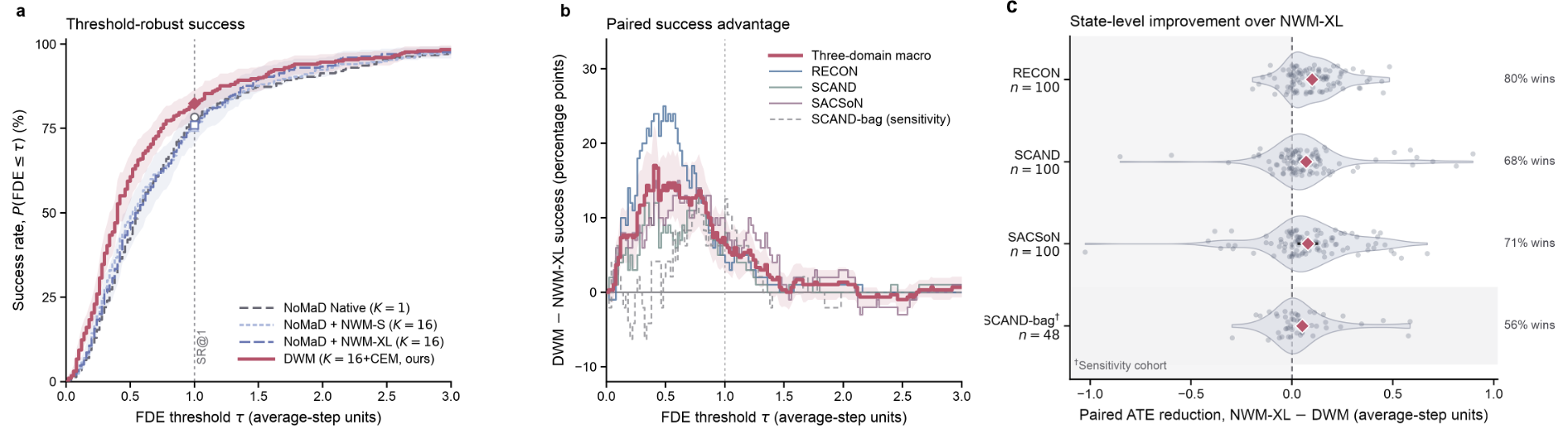}
\caption{Robustness across endpoint tolerances and evaluation states.
\textbf{(a)} Equal-domain macro endpoint agreement,
$P(\mathrm{FDE}\leq\tau)$; the vertical line marks SR@1.
\textbf{(b)} Paired endpoint-agreement advantage over NoMaD+NWM-XL,
with macro and domain-level estimates.
\textbf{(c)} Per-state ATE reduction relative to NoMaD+NWM-XL.
Grey points, violins, and diamonds show observations, distributions,
and means, respectively. Positive values favor LiteNWM; percentages
report per-state win rates.
Shading and horizontal bars indicate 95\% trajectory-cluster bootstrap
CIs from 20,000 resamples; bands in \textbf{(a,b)} are pointwise.
Each primary domain contains 100 states. SCAND-bag contains 48 states
from six clusters and is a descriptive sensitivity cohort excluded
from the macro average. DWM denotes the historical LiteNWM
configuration; success measures offline endpoint agreement.}
\label{fig:app_threshold}
\end{figure*}

Figure~\ref{fig:app_threshold} examines how endpoint advantage varies with tolerance and how frequently ATE improves across states. 
This historical evaluation contains 100 matched states per primary domain and retains its original configuration and candidate pools. 

At $\tau=1$, the historical macro endpoint-agreement advantage over NoMaD+NWM-XL is 6.33 percentage points.
Panels (a,b) show the corresponding cumulative endpoint distribution and paired difference across tolerances. The difference contracts as both methods approach saturation at looser thresholds, reaching
$-1$ percentage point at some points on the plotted grid.

Panel (c) shows that lower average error is accompanied by improvement on most states: 80\% on RECON, 68\% on SCAND, and 71\% on SACSoN.
The corresponding loss rates are 19\%, 32\%, and 29\%, with 1\% ties on RECON.

\Needspace{7\baselineskip}
\section{Acceptance Gate and Ensemble Diagnostics}
\label{app:gate}

The gate and ensemble diagnostics use the historical CEM3 configuration.
A replacement is harmful if either ATE or $J$ increases relative to
the incumbent, with tolerance $10^{-7}$~m.
The gate reduces the harmful replacement rate from 23.00\% to
9.92\%, a reduction of 13.08 percentage points
(95\% CI: $[10.22,\,16.16]$), while reducing replacement coverage
from 100\% to 64.83\% (Table~\ref{tab:app_gate}).
No-gate minus gated macro ATE is
$-0.00409$ (95\% CI: $[-0.00912,\,0.00076]$);
the interval includes zero.

\begin{table}[!t]
\centering
\LiteAppendixTableCaption{tab:app_gate}{
Historical CEM3 gate diagnostic over 1,200 paired state--seed
observations. Coverage is the replacement rate.
Coverage and harmful replacement rates are computed over
all 1,200 decisions.}
\DWMUniformTable{\linewidth}{\small}{5pt}{3}{l|cc}{
  \hline
  Configuration & Coverage (\%) & Harmful replacement (\%) \\
  \hline
  Gated   & 64.83  & \textbf{9.92} \\
  No gate & 100.00 & 23.00 \\
  \hline
}
\end{table}

Table~\ref{tab:app_ensemble} varies one ensemble at a time:
single-predictor runs retain three scorers, and single-scorer runs
retain three predictors. In the reported point estimates, the full
scorer ensemble has lower ATE than each individual scorer, while the
full predictor ensemble has higher ATE than each single-predictor
configuration. None of the six contrasts survives Holm correction.

\begin{table}[!t]
\centering
\LiteAppendixTableCaption{tab:app_ensemble}{
Historical ensemble-member diagnostic.
$\Delta$ATE is full ensemble minus single member; negative favors full.
CIs are unadjusted, and $p_{\mathrm{H}}$ uses six-test Holm correction.}
\DWMUniformTable{\linewidth}{\scriptsize}{3pt}{4}{l|ccc}{
  \hline
  Single member & $\Delta$ATE & 95\% CI & $p_{\mathrm{H}}$ \\
  \hline
  Predictor 1701 & $+0.00149$ & $[-0.00251,0.00537]$ & 0.9407 \\
  Predictor 1702 & $+0.00258$ & $[-0.00109,0.00643]$ & 0.7808 \\
  Predictor 1703 & $+0.00070$ & $[-0.00334,0.00481]$ & 0.9407 \\
  \hline
  Scorer 1701 & $-0.00735$ & $[-0.01349,-0.00159]$ & 0.0929 \\
  Scorer 1702 & $-0.00246$ & $[-0.00713,0.00216]$ & 0.8946 \\
  Scorer 1703 & $-0.00317$ & $[-0.00758,0.00109]$ & 0.7808 \\
  \hline
}
\end{table}

\section{Resource Measurement Details}
\label{app:runtime}

\begin{table}[!t]

\LiteAppendixTableCaption{tab:app_runtime}{
Runtime and resources. Values are repeat means, except peak VRAM.
Bold identifies the default LiteNWM configuration.
Measurement campaigns are described in the text.}
\DWMUniformTable{\linewidth}{\scriptsize}{3.5pt}{5}{l|rrrr}{
  \hline
  Configuration & Runs & Wall (s) & Energy (Wh) & VRAM (MiB) \\
  \hline
  NoMaD+NWM-S  & 3 & 11.653 & 1.5929 & 8,344 \\
  NoMaD+NWM-XL & 3 & 88.449 & 14.0695 & 13,776 \\
  \hline
  \textbf{LiteNWM, CEM1} & 4
    & \textbf{0.691} & \textbf{0.0277} & \textbf{2,459} \\
  LiteNWM, CEM3 & 4 & 1.179 & 0.0490 & 2,459 \\
  \hline
}
\end{table}

We measure wall time, GPU energy, and peak VRAM on an RTX~5090 by using RECON-100 with $H=4$ and $K=16$ (Table~\ref{tab:app_runtime}).
The NWM rerankers use three interleaved repeats, and the CEM-depth comparison uses four balanced repeats.
NWM totals combine the measured proposal-generation and reranking stages, with peak memory taken across stages.

LiteNWM wall time covers the complete pipeline specified in Sec.~\ref{subsec:efficiency}, including preprocessing and selection.
Its outer-process totals also include initialization and 30 warm-up states, divided by the 100 evaluated states.
Therefore, the reported values amortized benchmark costs per state. 
GPU energy is sampled over the measurement interval and excludes whole-robot consumption.

\makeatletter
\setlength{\@dblfptop}{0pt}
\setlength{\@dblfpbot}{0pt plus 1fil}
\makeatother

\begin{figure*}[!t]
    \centering

    \includegraphics[width=0.93\textwidth]{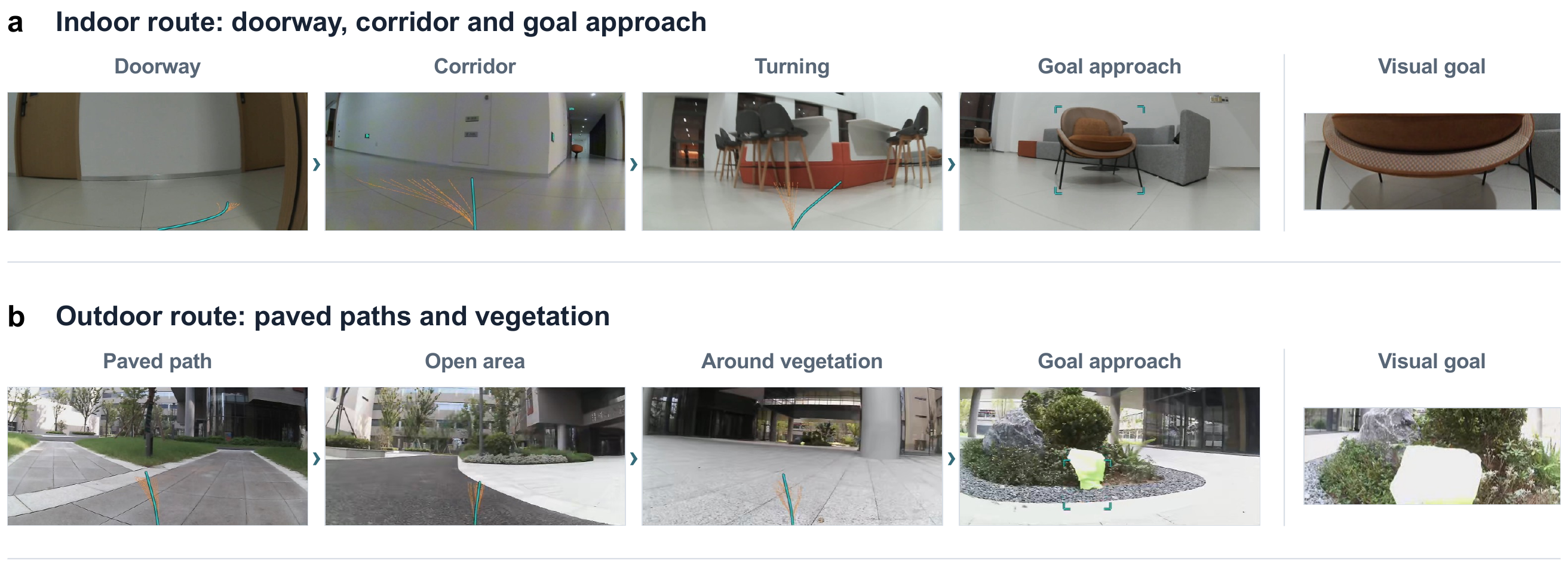}
    \caption{Indoor and outdoor route progression.
    \textbf{(a)} Doorway traversal, corridor navigation,
    turning, and approach to a chair goal.
    \textbf{(b)} Outdoor navigation across paved areas and beside
    vegetation toward the visual target.
    Four observations in each row follow video order from
    left to right, followed by a cropped visual-goal reference.
    Trajectory overlays reproduce the source-video visualization.
    Frame spacing does not represent uniform elapsed time.}
    \label{fig:app_deployment_routes}

    \par\vspace{4pt}

    \includegraphics[width=0.93\textwidth]{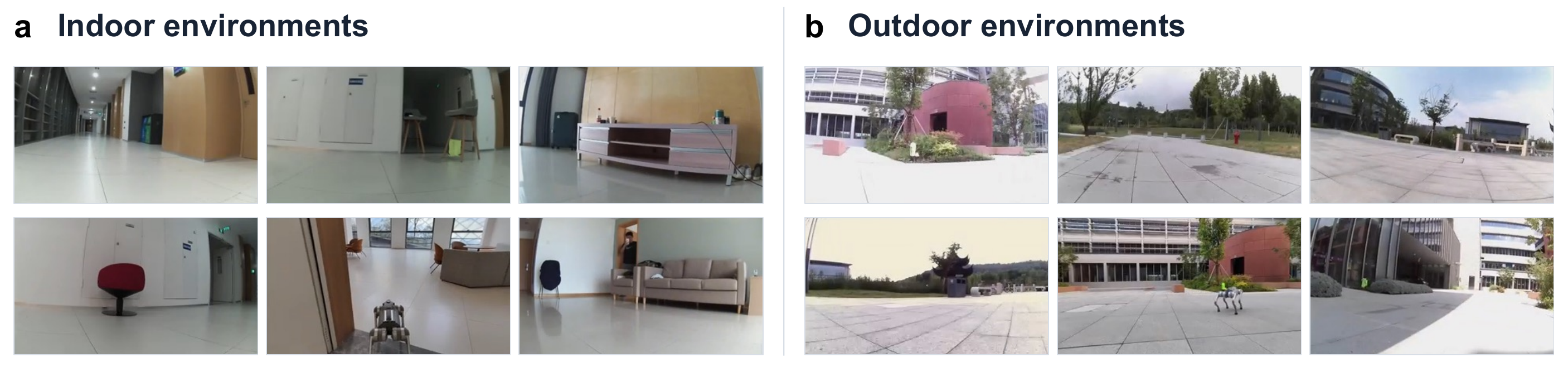}
    \caption{Indoor and outdoor deployment examples.
    \textbf{(a)} Corridors and furnished indoor spaces with
    varied objects and layouts.
    \textbf{(b)} Paved outdoor spaces, vegetation, and
    building-side passages under different illumination.
    Each panel contains six selected video views, including
    robot-camera and external views. Tiles illustrate visual
    diversity and do not denote independent environments
    or trials.}
    \label{fig:app_deployment_environments}
\end{figure*}

\section{Qualitative Real-World Deployment}
\label{app:deployment_cases}

\subsection{Route Progression}

Figure~\ref{fig:app_deployment_routes} supplies ordered observations
for the routes shown in Fig.~\ref{fig:real_world}(a). The indoor
sequence connects doorway traversal and corridor progression to a
turn beside furniture and approach to the chair goal. The outdoor
sequence connects paved-path travel, an open area, passage beside
vegetation, and approach to the visual target.
These excerpts illustrate continued route progression through
successive local motions under the fixed prediction horizon.

\subsection{Deployment Conditions}

Figure~\ref{fig:app_deployment_environments} adds views of the
conditions encountered during deployment: enclosed corridors and
furnished indoor spaces, as well as open paved areas, vegetated
borders, and building-side passages. The selected indoor and outdoor
views show changes in surrounding geometry, target appearance, and
sunlit or shaded conditions. Episode-level outcomes are reported in
Table~\ref{tab:real_world_results}.

\subsection{Onboard Deployment}
\label{app:onboard_deployment}

\sys runs on the onboard Jetson Orin of a Unitree Go2 EDU,
with all models frozen. Onboard odometry supports motion feedback
and, in the extended-route demonstrations, route progression.

The small aggregate error reduction from CEM1 to CEM3 in
Table~\ref{tab:app_search} supports CEM1 as the offline default.
Onboard deployment skips CEM when the initial pool meets the
acceptance conditions; otherwise, it performs up to three
refinement rounds, stopping once an eligible challenger is
available. Acceptance uses the learned gate and physical and
proposal-support checks in Sec.~\ref{subsec:selection}, together
with execution checks. Across 311 logged planning calls,
four used two CEM rounds and 19 used three rounds.

For the extended-route demonstrations, a pre-mapped visual route
supplies a predefined sequence of intermediate goal images.
Route memory manages progress and automatically switches the
active visual goal, while the same frozen \sys evaluator performs
local trajectory selection at $H=4$.

\subsection{Onboard Planning Latency}
\label{app:onboard_latency}

Table~\ref{tab:app_onboard_latency} reports internal planning
latency from the robot logs. For \sys, this includes shared
visual encoding, proposal generation, future evaluation, scoring,
conditional refinement or recovery, and final selection.
The interval excludes model initialization and robot motion.
All completed planning calls from the 30 \sys trials in
Table~\ref{tab:real_world_results} are included, including calls
that abstain from execution. Statistics pool decisions across
trials; P95 uses the nearest-rank definition.
NoMaD latency is measured from 302 planning calls recorded
across 30 real-world runs.

\begin{table}[!t]
\centering
\LiteAppendixTableCaption{tab:app_onboard_latency}{
Onboard internal planning latency on Jetson Orin, in seconds
per completed planning call. NoMaD+NWM-S uses $K=16$, $H=4$,
one rollout repeat, and DDIM-30 across two runs, and is excluded
from the SR/SPL comparison.}
\DWMUniformTable{\linewidth}{\scriptsize}{3pt}{5}{l|rrrr}{
  \hline
  Configuration & Calls & Mean (s) & Median (s) & P95 (s) \\
  \hline
  NoMaD       & 302 & 0.651   & 0.562   & 1.347   \\
  \sys        & 311 & 2.228   & 2.085   & 3.883   \\
  NoMaD+NWM-S & 22  & 235.038 & 230.834 & 252.300 \\
  \hline
}
\end{table}

With $K=16$, three rollout repeats, and DDIM-100,
NWM-S reranking on the same Jetson Orin takes 692.7~s at
$H=4$ and 1381.2~s at $H=8$, each measured on one saved
observation--goal state. The $H=8$ profile uses the trajectory
horizon of the original NWM~\cite{nwm}.
These measurements cover reranking only.
Onboard results are reported separately from the RTX~5090
benchmark in Sec.~\ref{subsec:efficiency} and
Appendix~\ref{app:runtime} because the hardware,
configurations, and timing boundaries differ.

\section{Discussion}
\label{app:Discussion}

Fixed-pool selection retains 94.75\% of the observed CEM1 ATE
reduction relative to NoMaD+NWM-XL (Table~\ref{tab:app_search}).
Increasing CEM depth from one to three rounds lowers macro ATE
by a further 0.64\%, while raising wall time by approximately
71\% in the RECON resource benchmark (Table~\ref{tab:app_runtime}).
CEM here refines policy proposals, whereas NWM~\cite{nwm} also
uses it for standalone planning. Comparing optimization gains
requires matched objectives and budgets; these experiments do not
distinguish limited search coverage from errors in evaluating
refined trajectories.

The frozen proposer receives no training feedback from the evaluator,
so its initial proposal distribution remains unchanged when \sys
identifies better trajectories. The four-step predictor evaluates
local consequences; longer routes rely on repeated replanning and
do not establish prediction accuracy beyond this horizon.
More aggressive score optimization could expose model errors,
particularly outside the proposal distribution.

We aim to train the trajectory generator with scorer and predictor
feedback. Inspired by DINO-WM~\cite{dinowm}, gradients of a smooth
incumbent-relative cost could pass through the scorer, predictor,
and candidate actions to a differentiable generator's parameters.
Evaluator weights, the incumbent, and comparison context would
initially remain fixed, with updates constrained by kinematics and
proximity to the proposal distribution. Evaluation must test observed
trajectory outcomes as well as predicted scores. Comparisons with CEM
and random expansion should match computational budgets, including
backward-pass costs. Distilling refined trajectories could reduce
repeated optimization during deployment.

Beyond the predefined intermediate goal images used in the
extended-route demonstrations, persistent memory and autonomous
hierarchical subgoal selection could support detours and partially
observed routes. Exploration and search-and-rescue additionally
require choosing destinations and discovering targets. A higher-level
planner could select subgoals using coverage, information gain, or
semantic target detection, while \sys evaluates local trajectories.
Training would require targets for exploration progress, discovery,
and traversability under mission time or energy constraints.
Evaluation would measure coverage, discovery rate, and successful
execution in unseen environments, sustained onboard latency,
and whole-robot energy.


\begin{thebibliography}{00}
\bibitem{targetnav}
Y. Zhu, R. Mottaghi, E. Kolve, J. J. Lim, A. Gupta, L. Fei-Fei,
and A. Farhadi,
Target-driven visual navigation in indoor scenes using deep reinforcement learning,'' in \emph{Proc. IEEE Int. Conf. Robot. Autom. (ICRA)}, 2017, pp. 3357--3364, doi: 10.1109/ICRA.2017.7989381. 
\bibitem{gnm} D. Shah, A. Sridhar, A. Bhorkar, N. Hirose, and S. Levine, GNM: A general navigation model to drive any robot,''
in \emph{Proc. IEEE Int. Conf. Robot. Autom. (ICRA)}, 2023.
\bibitem{vint}
D. Shah, A. Sridhar, N. Dashora, K. Stachowicz, K. Black,
N. Hirose, and S. Levine,
ViNT: A foundation model for visual navigation,'' in \emph{Proc. Conf. Robot Learning (CoRL)}, 2023. 
\bibitem{nomad} A. Sridhar, D. Shah, C. Glossop, and S. Levine, NoMaD: Goal masked diffusion policies for navigation and exploration,''
in \emph{Proc. IEEE Int. Conf. Robot. Autom. (ICRA)}, 2024.
\bibitem{nwm}
A. Bar, G. Zhou, D. Tran, T. Darrell, and Y. LeCun,
Navigation world models,'' in \emph{Proc. IEEE/CVF Conf. Comput. Vis. Pattern Recognit. (CVPR)}, pp. 15791--15801, 2025. 
\bibitem{navdp} W. Cai et al., NavDP: Learning sim-to-real navigation diffusion policy with privileged
information guidance,''
arXiv preprint arXiv:2505.08712, 2025.
\bibitem{vilint}
L. Dezons, Q. Picard, R. Marsal, F. Goulette, and D. Filliat,
Multimodal embodiment-aware navigation transformer,'' arXiv preprint arXiv:2604.19267, 2026. 
\bibitem{pathdreamer} J. Y. Koh, H. Lee, Y. Yang, J. Baldridge, and P. Anderson, Pathdreamer: A world model for indoor navigation,''
in \emph{Proc. IEEE/CVF Int. Conf. Comput. Vis. (ICCV)},
pp. 14738--14748, 2021.
\bibitem{planet}
D. Hafner, T. Lillicrap, I. Fischer, R. Villegas, D. Ha,
H. Lee, and J. Davidson,
Learning latent dynamics for planning from pixels,'' in \emph{Proc. Int. Conf. Mach. Learn. (ICML)}, 2019. 
\bibitem{tdmpc2} N. Hansen, H. Su, and X. Wang, TD-MPC2: Scalable, robust world models for continuous control,''
in \emph{Proc. Int. Conf. Learn. Represent. (ICLR)}, 2024.
\bibitem{vjepa2}
M. Assran et al.,
V-JEPA 2: Self-supervised video models enable understanding, prediction and planning,'' arXiv preprint arXiv:2506.09985, 2025. 
\bibitem{dinowm} G. Zhou, H. Pan, Y. LeCun, and L. Pinto, DINO-WM: World models on pre-trained visual features enable
zero-shot planning,''
in \emph{Proc. ICML}, 2025.
\bibitem{dinoworld}
F. Baldassarre et al.,
Back to the features: DINO as a foundation for video world models,'' arXiv preprint arXiv:2507.19468, 2025. 
\bibitem{df3} J. Chen et al., DF$^{3}$: World modeling via decoder-free feature forecasting
in autonomous navigation,''
arXiv preprint arXiv:2608.02428, 2026.
\bibitem{relnwm}
Z. Zhang, H. Zhang, K. Huang, C. Shi, and H. Lu,
``Efficient image-goal navigation with representative latent world model,''
arXiv preprint arXiv:2511.11011, 2025,
doi: 10.48550/arXiv.2511.11011.
\bibitem{wamnav}
N. Yang et al.,
WAM-Nav: Asymmetric latent world-action modeling for unified visual navigation,'' arXiv preprint arXiv:2606.04907, 2026. 
\bibitem{navwam} D. Azuma et al., NavWAM: A navigation world action model for goal-conditioned
visual navigation,''
arXiv preprint arXiv:2606.13494, 2026.
\bibitem{diffusionpolicy}
C. Chi, Z. Xu, S. Feng, E. Cousineau, Y. Du, B. Burchfiel,
R. Tedrake, and S. Song,
Diffusion policy: Visuomotor policy learning via action diffusion,'' in \emph{Proc. Robot. Sci. Syst. (RSS)}, 2023, doi: 10.15607/RSS.2023.XIX.026. \bibitem{vjepa} A. Bardes et al., Revisiting feature prediction for learning visual representations from
video,'' arXiv preprint arXiv:2404.08471, 2024.
\bibitem{dinov3}
O. Sim{\'e}oni et al.,
DINOv3,'' arXiv preprint arXiv:2508.10104, 2025. 
\bibitem{recon} D. Shah, B. Eysenbach, N. Rhinehart, and S. Levine, Rapid exploration for open-world navigation with latent goal models,''
in \emph{Proc. Conf. Robot Learning (CoRL)}, PMLR, vol. 164,
pp. 674--684, 2022.
\bibitem{scand}
H. Karnan et al.,
Socially CompliAnt Navigation Dataset (SCAND): A large-scale dataset of demonstrations for social navigation,'' \emph{IEEE Robot. Autom. Lett.}, vol. 7, no. 4, pp. 11807--11814, 2022, doi: 10.1109/LRA.2022.3184025. \bibitem{sacson} N. Hirose, D. Shah, A. Sridhar, and S. Levine, SACSoN: Scalable autonomous control for social navigation,''
\emph{IEEE Robot. Autom. Lett.}, 2024,
doi: 10.1109/LRA.2023.3329626.
\bibitem{deepensemble}
B. Lakshminarayanan, A. Pritzel, and C. Blundell,
Simple and scalable predictive uncertainty estimation using deep ensembles,'' in \emph{Adv. Neural Inf. Process. Syst.}, vol. 30, pp. 6402--6413, 2017. 
\bibitem{cem} M. Kobilarov, Cross-entropy motion planning,''
\emph{Int. J. Robot. Res.}, vol. 31, no. 7, pp. 855--871, 2012,
doi: 10.1177/0278364912444543.
\bibitem{mbra}
N. Hirose, L. Ignatova, K. Stachowicz, C. Glossop, S. Levine,
and D. Shah,
``Learning to drive anywhere with model-based reannotation,''
arXiv preprint arXiv:2505.05592, 2025.
\end{thebibliography}
\end{document}